# Spectral Graph Cut from a Filtering Point of View

Chengxi Ye, Mingli Song, Yuxu Lin, Chun Chen and David W. Jacobs

**Abstract**—Spectral graph theory is well known and widely used in computer vision. In this paper, we analyze image segmentation algorithms that are based on spectral graph theory, e.g., normalized cut, and show that there is a natural connection between spectural graph theory based image segmentationand and edge preserving filtering. Based on this connection we show that the normalized cut algorithm is equivalent to repeated iterations of bilateral filtering. Then, using this equivalence we present and implement a fast normalized cut algorithm for image segmentation. Experiments show that our implementation can solve the original optimization problem in the normalized cut algorithm 10 to 100 times faster. Furthermore, we present a new algorithm called conditioned normalized cut for image segmentation that can easily incorporate color image patches and demonstrate how this segmentation problem can be solved with edge preserving filtering.

## 1 Introduction

Image segmentation is a fundamental task in machine vision. A wide range of computational vision problems benefit from making use of segmented images. For instance, intermediate-level vision problems such as stereo and motion estimation require an appropriate region of support for correspondence operations. Spatially non-uniform regions of support can be identified using segmentation techniques. Higher-level problems such as recognition and image indexing also make use of segmentation results in matching, to address problems such as figure-ground separation and recognition by parts[1].

In graph cut based image segmentation, the image is modeled as a weighted undirected graph. Nodes are pixels and weighted edges between the nodes measure the pair-wise similarity of the pixels. Segmentation is formulated as the problem of finding the optimal cut in this weighted graph. However, solving this optimization problemis generally difficult. Hence, a variety of approximations have been proposed [2-4] to tackle this problem.

Spectral graph theory based cut algorithms are a popular approach that tackles the problem mentioned above by analyzing the weight/affinity matrix. Intuitively, similar elements should be grouped in one cluster, and we can formulate this as an optimization problem subject to reasonable constraints (e.g. the number of elements or the sum of weights within each group should be balanced). Usually solving the constrained optimization problem involves finding the eigenvectors corresponding to specific eigenvalues (see section 2). Spectral graph theory based segmentation can be regarded as a generalization of Markov random field (MRF) based segmentation techniques [5, 6] since the affinity matrix allows dependence of non-adjacent pixels.

The normalized cut algorithm [4] is a typical spectral graph theory based graph cut technique that has achieved a remarkable position in recent years. This algorithm tries to cut the graph into balanced segments (normalized by the sum of the edge weights in each segment). Normalized cut has been widely used in image segmentation due to its nice algebraic property: the problem is reduced to finding eigenvectors of a Laplacian matrix. However, a major limitation of the method is its high computational demands when the image becomes large (since usually the graph radius needs to be large). Traditionally solving this problem is accelerated in several ways: : the simplest way is to sample the pixels in the connection radius, so the affinity matrix is sparser[4, 7], this approach is implemented in the original normalized cut. Another intuitive way is to use multiscale analysis [8] to downsample the image so a small graph radius in the coarse scale includes distant affinities in the original scale. Interscale consistency is promised by carefully putting all the scales together into a larger problem and by adding a new consistency constraint among all the scales. The other approach introduces prior information to constrain the solution and makes use of more efficient optimizations [9, 10]. In general, these methods either reformulate normalized cut as another optimization problem or rely on prior information to make the problem simpler.

In contrast with the high computational cost of the spectral graph theory based graph cut, in image filtering, researchers have successfully found numerous ways to speed up the edge preserving filtering operation [11, 12]. The bilateral filter [13, 14] is perhaps the most typical effort among them. The bilateral filter is widely used in many image enhancement problems such as denoising [13, 15], tone mapping [16], flash/non-flash photo fusion [17, 18] and stylization [19]. For each pixel, the bilateral filter calculates the weighted average of nearby pixels within radius *r*. The weight mask is constructed by putting two criteria together: nearby pixels have higher weight; pixels with similar intensity have higher weight. These considerations lead to multiplying a range kernel by a domain kernel as the weight mask ([14] or see Section 2). Over the years, there have been many advances [20-22] in accelerating the bilateral filter. Recent breakthroughs have shown real time applications with complexity reduced to $O(N)$ [23, 24], regardless of the domain and range kernel size. There have been many recent efforts that generalize edge preserving filters [13, 25-27].

In this paper, by further investigating and understanding the normalized cut and the bilateral filter, we find a natural connection between them, though they look totally different from a conventional view. This connection enables us to build a novel fast bilateral nCut algorithm, that benefits from the recent work to speed up the bilateral filter. In contrast with previous accelerations, we solve the *original* optimization problem directly. To the best of our knowledge, this is the first work that explicitly builds the theoretical connection between the spectral graph theory based graph cut and edge preserving filtering. Furthermore, we extend the conditional random fields [28] in a non-local way and propose a new conditional normalized cut model. We show how this model can be solved with filtering techniques.

The rest of this paper is organized as follows. We first investigate and discuss the spectral graph cut and edge preserving filter in more depth in Section 2. We discuss implementation details in Section 3. Evaluation and experiments are carried out in Section 4. Finally, we conclude in Section 5.

## 2 Spectral Graph Cut and Edge Preserving Filter

Though spectral graph cuts and edge preserving filtering have been intensively investigated, fewer efforts have been made to explore their connection. In this section we show the two fields are actually closely related and that spectral graph cuts can benefit from this connection to speed up the image segmentation operation greatly. As representatives for spectral graph cut and edge preserving filter, normalized cut and the bilateral filter are theoretically explored and discussed in this section. We then propose a new model called conditioned normalized cut and explain its connection with edge preserving filtering.

### 2.1 Normalized Cut and Bilateral Filtering

In the normalized cut algorithm, the weight matrix between pixels is constructed to measure the similarity between the pixels. With such a weight matrix, we can calculate a cut by minimizing the cost $\sum_{i,j} w_{ij}(y_i - y_j)^2$, where $w_{ij}$ is the $(i, j)$ element of the weight matrix $W$. $w_{ij}$ measures the similarity of the $i$ th pixel and the $j$ th pixel, and $y$ represents the segmentation labels. The cost function can also be written in a matrix form: $y^T(D-W)y$ ($D$ is a diagonal matrix that records the row sum of $W$). The normalized cut algorithm uses the association as normalization and is formulated as minimizing the Rayleigh quotient[4], or finding an embedding into 1-d space [29]: $\min_y \frac{y^T(D-W)y}{y^T Dy}$ subject to $y^T D1 = 0$ .While a normalized cut requires that $y$ have binary values, this constraint is first relaxed, so that we find a continuous solution. Postprocessing methods are then applied to find a binary solution near the continuous one. We do not discuss this final aspect of the problem in detail in this paper, as similar techniques can be applied to all methods we discuss.

Solving this continuous problem is equivalent to solving the generalized eigenvalue system: $(D-W)y = \lambda Dy$ , where $\lambda$ is the second smallest eigenvalue of the system.

Commonly this system is solved by rewriting the above equation as:

$$[D^{-\frac{1}{2}}(D-W)D^{-\frac{1}{2}}][D^{\frac{1}{2}}y] = \lambda[D^{\frac{1}{2}}y], \quad (1)$$

due to the numerical advantages of a symmetric system. In order to measure similarity, the weight matrix was suggested with the criteria that neighboring pixels with similar intensity should have higher weight. Explicitly, one simple and natural strategy [4] to compute the weight for each pixel is:

$$w_{ij} = \exp(-\frac{(x_i-x_j)^2}{2\sigma_X{}^2})\exp(-\frac{(I_i-I_j)^2}{2\sigma_I{}^2}). \quad (2)$$

We will consider this weight, but our theory extends readily to other choices of weights.

On the other hand, the bilateral filter smoothes an image in an edge preserving way, also by assigning higher weights to neighboring pixels with similar intensity and then by taking the weighted average. The standard bilateral filter [14] is expressed as:

$$I_i = \frac{\sum_{j\in N_i} w_{ij} I_j}{\sum_{j\in N_i} w_{ij}} = \frac{\sum_{j\in N_i} \exp(-\frac{(x_i-x_j)^2}{2\sigma_X{}^2})[\exp(-\frac{(I_i-I_j)^2}{2\sigma_I{}^2})I_j]}{\sum_{j\in N_i} \exp(-\frac{(x_i-x_j)^2}{2\sigma_X{}^2})[\exp(-\frac{(I_i-I_j)^2}{2\sigma_I{}^2})]}, \quad (3)$$

which uses the same weights as above.

### 2.2 The Underlying Connection Between Normalized Cut And Bilateral Filter

From the above description of normalized cut and the bilateral filter, a natural connection between them can be derived.

**Theorem:** With the weight matrix in **(2)**[4, 14], normalized cut is equivalent to finding the second largest eigenvalue and the corresponding eigenvector of the bilateral filter.

Proof: Rather than transforming the generalized eigenvector problem into a symmetric form **(1)**, we take the inverse of $D$, and solve the optimization problem as $(D^{-1}W)y = (1-\lambda)y$. Since $D$ is diagonal and records the row sum of $W$, left multiplication using $D^{-1}W$ is eqivalent to calculating the weighted average of $y$ with weight sum 1 for each pixel. That is, multiplication by $D^{-1}W$ is exactly the application of the bilateral filter. $\lambda$ is the second smallest eigenvalue of the original system, so $1-\lambda$ corresponds to the second largest eigenvalue of the bilateral filter. Note that the largest eigenvalue of $D^{-1}W$ is the constant vector. Therefore, repeated application of the bilateral filter corresponds to the power method for finding eigenvectors, and will produce an image that contains a DC component plus the other eigenvectors associated with normalized cut.□

The above theorem gives us a new intuitive interpretation of the normalized cut: the eigenvectors calculated in the algorithm are the stable distributions under edge preserving filtering.

This connection also hints at a new way to accelerate the eigenvector calculation. In order to solve **(1),** traditional solvers require hundreds of iterations of left multiplication with $D^{-\frac{1}{2}}(D-W)D^{-\frac{1}{2}}$ in the power method, which becomes slow and memory intensive when the radius becomes large. Now we replace this by calculating left multiplications using $D^{-1}W$, which can be accelerated using fast bilateral filtering, thanks to recent advances[16, 20, 22-24].

## 2.3 Other Pixel Based Spectral Graph Cuts

Simple modifications in accelerating the bilateral filter tell us similar accelerations can be extended to other cuts [4], including average association cut $Wx = \lambda x$ and average cut $(D-W)x = \lambda x$, since the calculations of $Wx$ and $Dx$ are usually carried out independently in bilateral filter accelerations [20, 24].

## 2.4 Conditional nCut: A Patch Based Normalized Cut

Based on the natural connection, we can borrow wisdom from the filtering field and enhance the efficacy of segmentation models [13, 26]. Since image patches provide a rich source of information compared with pixel intensity, the conditional random fields (CRF) model takes advantage of this information [28] to make inferences. In segmentation, the affinity of two adjacent pixels is usually measured by a probability conditioned on the local patches. Inference is made iteratively on this resulting Markov random field, but is usually extremely slow.

Here we propose a new patch based model *without* the restriction of the Markov property. We measure the affinity of two pixels by the similarity their neighborhoods, i.e. the local patches. The segmentation is performed under the criterion that pixels with similar patches shall be grouped into the same cluster.

Thus the affinity of two pixels is conditioned on their surrounding pixels and can be measured by:

$$w_{ij} = \exp(-\tfrac{(x_i - x_j)^2}{2\sigma_X^2})\exp(-\tfrac{\|N_i - N_j\|^2}{2\sigma_N^2}) \qquad \textbf{(4)}$$

where $N_i$, $N_j$ are the patches with $i$, $j$ as center pixels, and the distance of two patches is simply the $l^2$ distance, or better, a gaussian weighted distance assigning higher weights to the center pixels. Note that it is straightforward to put color information into this model.

We call the new cut problem with the weight as in **(4)** the conditional normalized cut problem.

**Corollary:** With the weight matrix in **(4)**, the conditioned normalized cut is equivalent to finding the second largest eigenvalue and the corresponding eigenvector of the nonlocal means filter[13].

Proof: by comparing the $w_{ij}$ with the weight in [13].

## 2.5 Preprocessing and Other Filtering Models

One can use any filtering technique to pre-smooth or stylize the input image [19, 25-27]. Another more recent idea widely used in segmentation is to take into consideration the intervening contours between two points when determining their affinity. Under this framework [30], propagation is penalized across contour lines. While we do not explicitly pursue this approach in this paper, this property can be enforced by using geodesic distance in filtering [26]. And in our conditioned normalized cut algorithm implementation, this geodesic distance based diffusion is used for fast blurring, which results in a hybrid geodesic/Euclidean segmentation.

# 3 Implementations

It is now natural to build novel algorithms to enhance the efficiency and efficacy of normalized cut benefiting from the fast filtering techniques. Based on the natural connection, the only thing we need change is the calculation in the eigensolver. The eigenvectors of sparse matrices are usually numerically solved by the power method, which calculates the dominant eigenvector of $A$ by iteratively multiplying a random vector $b_0$: $b_{k+1} = \frac{Ab_k}{\|Ab_k\|} = \frac{A^{k+1}b_0}{\|A^{k+1}b_0\|}$. Decomposing $b_0$ into the basis of eigenvectors of $A$ tells us the coefficients of the non-dominant eigenvectors decay in the iterations and the series converges to the dominant eigenvector. The non-dominant eigenvectors are then calculated in a subspace. A well established adapted power method is the Lanczos algorithm[31], which solves multiple eigenvectors together efficiently. As in the original normalized cut, we use this algorithm to calculate the eigenvectors of $D^{-1}W$, but replace the matrix multiplication in each iteration by fast (joint) filtering. The out-put eigenvectors are then discretized to obtain the segmentation results.

One noteworthy point is that $D^{-1}W$ encodes the weight calculated from the initial image, which is fixed in all iterations. Because of this, variants of the bilateral filter/ non-local means filter, i.e. the joint bilateral/non-local means filters [17, 18], are needed in this acceleration. Joint filtering uses weights calculated from another 'joint' image. We fix this joint image as the initial image to be segmented, so the weight is calculated from this image. Another interesting thing, as noted in [8], is that the eigenvector solving process is similar to anisotropic diffusion -- it takes longer to propagate the grouping cues with small connection radius. Thus larger radius usually suggests *better* segmentation results and converges in *fewer* iterations. This also hints at why a spectral graph cut approach can be preferable to the MRF/CRF approach. Since the complexity of most recent filtering techniques is no more expensive with increased radius, using a larger radius leads to fewer iterations and further acceleration. Last, blurring an image with a low pass filter leads to a band limited signal that can be well approximated from low frequency, following the Shannon sampling theorem. Thus the blurring (which is needed in accelerating the bilateral filter**(3)**) can be carried out in a coarse scale [20] and thus the calculation can be better accelerated with a larger connection radius.

## 4 Experimental Results and Discussions

Though there are various ways to accelerate the normalized cut, almost all of them formulate the problem into different optimizations. In contrast, our way solves the *original* optimization, it is general and could be combined with related techniques for further improvements. Thus in this section we only carry out comparisons with the standard normalized cut [4] and only change the settings in the eigensolver. In our current implementation, we use the latest standard normalized cut implementation from the authors and modified the eigensolver using a bilateral filter implementation taken from a classic benchmark work [20]. Recent accelerations of the bilateral filter have produced indistinguishable approximations with PSNR>40. It is noteworthy here that our current implementation shows the feasibility of introducing fast filtering to help segmentation; approximation analysis can be found in each of the related works. Also, specific speedups will increase with the latest developments of fast filtering techniques [23-25].

In the first example (Figure 1) we show a comparison on a 300x300 image. We used the standard normalized cut algorithm with sampling radius 15 and this is fixed in all the experiments. The brute force normalized cut with no pixel sampling used 326 filterings and took 58 seconds on a laptop computer with an Intel i7 3610QM CPU. As introduced in the original work [4], a default sampling ratio of 0.3 can be used to keep the weight matrix sparse and improve the speed. With this setting the segmentation using the standard normalized cut took 18 seconds to solve the eigensystem with 295 matrix multiplications. On the same platform, we used 1/32 image height as domain sigma for the bilateral filter; 141 joint bilateral filterings were used and it took 1.5 second to solve the eigensystem. Since we are solving the original normalized cut problem, our segmentation result resembles the original normalized cut results, this fact is demonstrated in Figure 1.

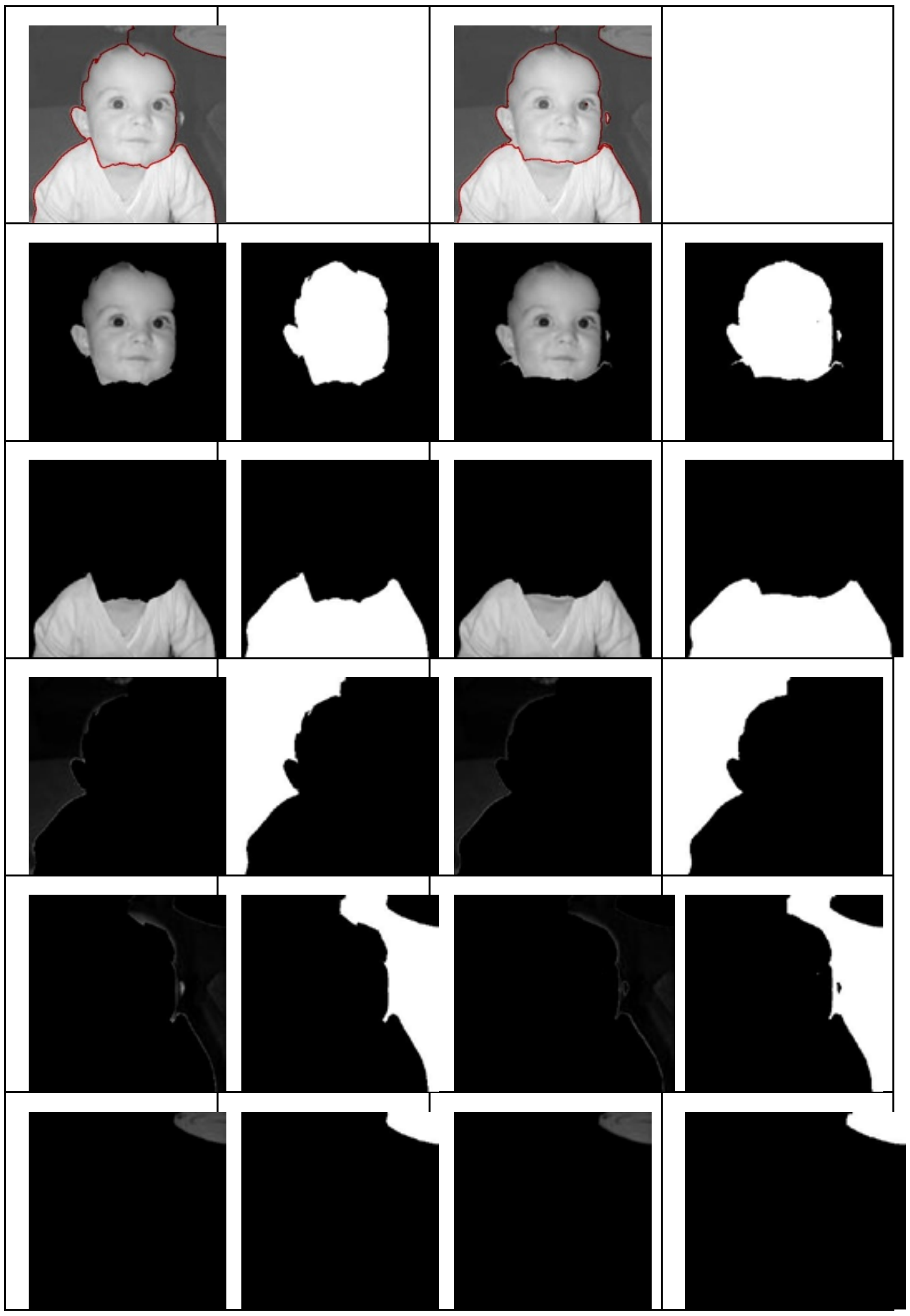

Figure 1: Comparison of the standard normalized cut and the bilateral normalized cut. First column: segmentation from the standard normalized cut. Second column: binary mask of each segment from the standard normalized cut. Third column: segmentation from the bilateral normalized cut. Fourth column: binary mask of each segment from the bilateral normalized cut.

We also notice as in [8] that a large domain sigma can lead to much better segmentation results. With fixed sampling radius at 15, in the second experiment (Figure 2) we found the segmentation was close to correct only when the flower image was resized to as small as 150x100 for the standard normalized cut algorithm (Figure 2, second column). At this size solving the eigensystem took 166 matrix multiplications and 5.3 seconds with no sam-pling or 133 matrix multiplications and 1.5 seconds with sampling ratio 0.3. At a scale of 300x200, the standard algorithm used 231 matrix multiplications and 30 seconds with no sampling or 198 matrix multiplications and took 8.5 seconds. The segmentation result was not accurate since the radius of 15 is not enough at this scale. In the bilateral filter setting we set 1/4 of the image height as the domain sigma for the image at 150x100, 300x200, 1500x1000 and 3000x2000 resolutions. The computation used an average of 16 bilateral filters and took 0.03, 0.12, 3.5 and 22 seconds. Memory consumption at 3000x2000 was ~700MB. At all resolutions fine segmentation was obtained with accurate cut at the boundary of the flower.

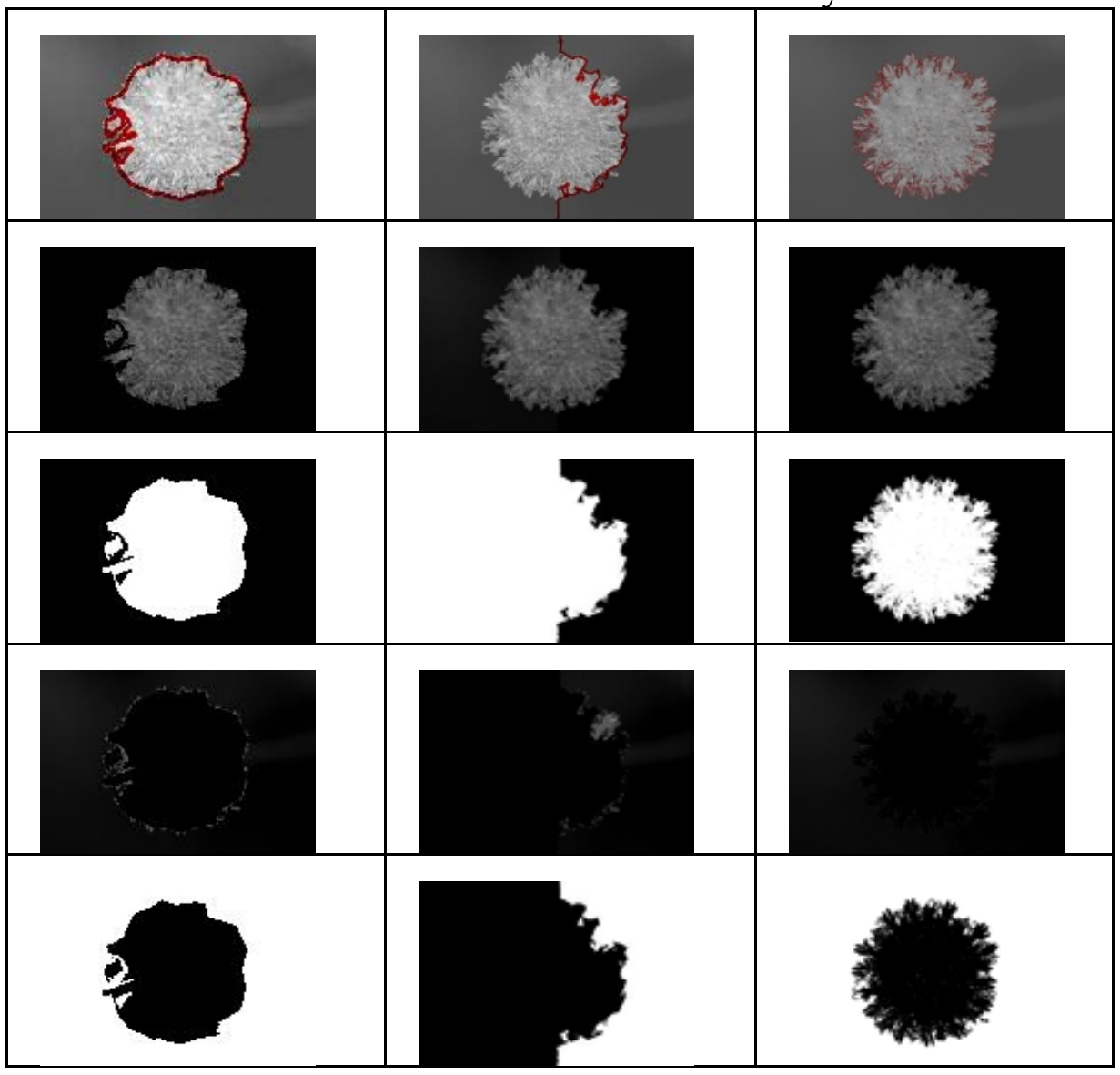

Figure 2: Comparison of the standard normalized cut and the bilateral normalized cut. First column: segmentation with the standard normalized cut at 150x100 resolution. Second column: segmentation with the standard normalized cut at 300x200 resolution. Third column: segmentation using the bilateral filter at 1500x1000 resolution. First row: the segmentations. Second row: the first segment. Third row: the binary mask corresponding to the first segment. Fourth row: the second segment. Fifth row: the binary mask of the second segment.

Finally, we show the conditioned normalized cut compared with the standard normalized cut. Both models perform similarly, but since the conditioned normalized cut can easily incorporate color and patch information, it makes more accurate segmentations especially for scenes with a complicated color distribution. We show segmentation results with 200x200 color/grayscale images, and use 5x5 patches in the conditional models. Our current implementation of non-local means filtering is taken from [25] and is written in Matlab. It takes ~10 seconds for the conditioned normalized cut algorithm to segment a color image, and ~ 5 seconds for a grayscale image which is on par with the standard normalized cut algorithm with sampling ratio 0.3. The bilateral filter accelerated normalized cut takes only 0.5 second.

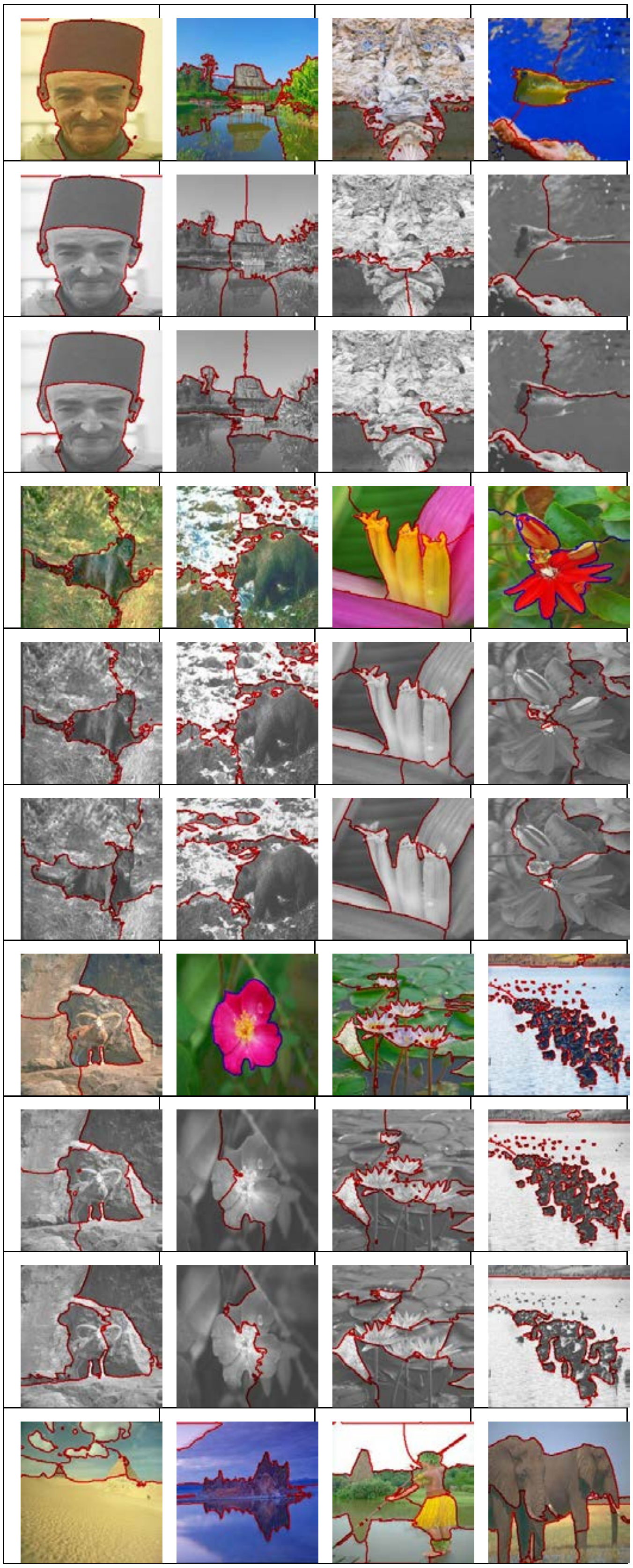

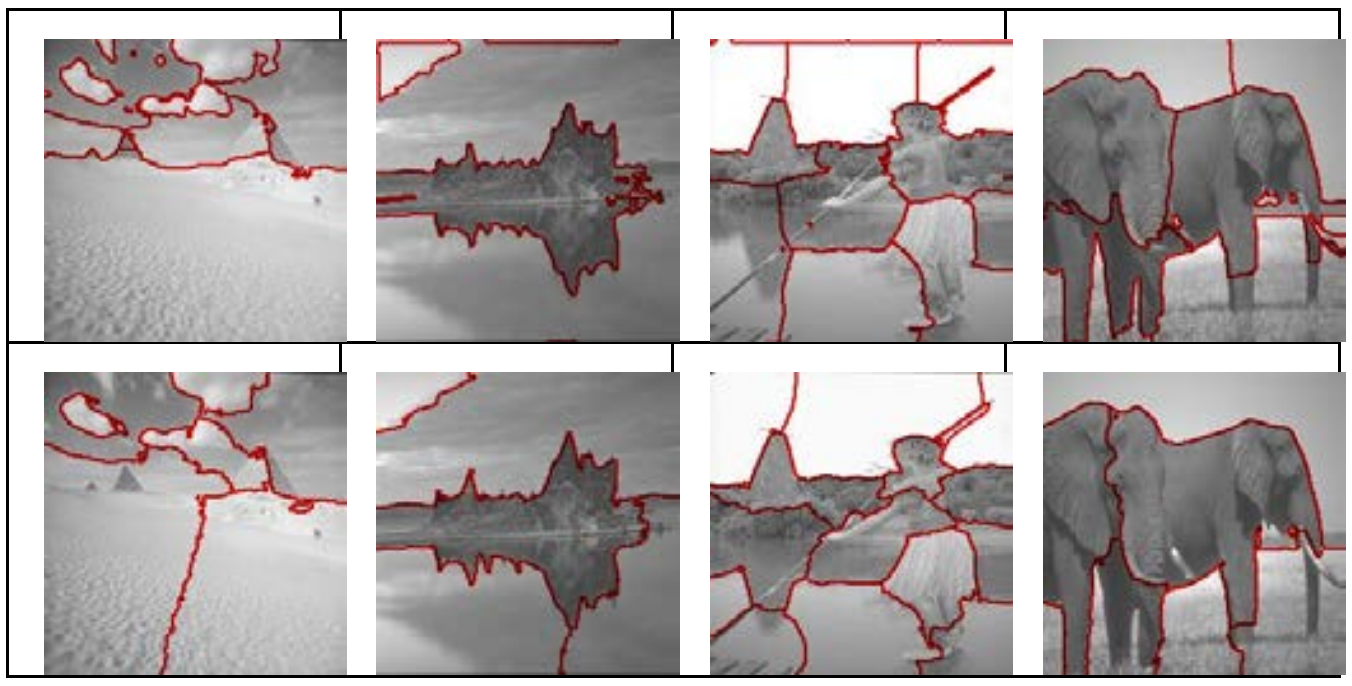

Figure 3: Comparisons of the conditioned and the standard normalized cut. Rows 1,4,7: segmentations with the conditioned normalized cut on color images. Rows 2, 5, 8: segmentations with the conditioned normalized cut on grayscale images. Rows 3,6,9: segmentations with the standard normalized cut.

## 5 Conclusion

Historically there has been some insightful thoughts majorly initiated by mathematicians about filtering and segmentation Section 1.3.2 in [32]. In this work we explicitely build the connection between spectral graph cut and edge preserving filtering. Based on the connection we borrow from the wisdom of recent breakthroughs in image filtering to accelerate graph cut and give a detailed explanation and implementation for the example of normalized cut. Currently we only show direct gains from this connection. It is expected that this approach can be combined with other advances in accelerating the spectral cut. Also we expect one could solve the eigensystem in a coarse-to-fine fashion which might lead to further improvements especially in large scale problems. The techniques in our work can also be extended to accelerate 3-d segmentation. Most importantly, with the connection we describe in this paper, novel ideas developed in one field may directly lead to new models in the other. We hope other researchers will benefit from this connection.

## Acknowledgment

The authors wish to thank A, B, C. This work was supported in part by a grant from XYZ.

## References

[1] Hoiem, D., Rother, C. and Winn, J. (2007) 3D LayoutCRF for Multi-View Object Class Recognition and Segmentation. Computer Vision and Pattern Recognition, 2007. CVPR '07. IEEE Conference on.

[2] Boykov, Y. and Funka-Lea, G. (2006) Graph Cuts and Efficient N-D Image Segmentation. *Int. J. Comput. Vision*. **70**(2) 109-131.

[3] Boykov, Y., Veksler, O. and Zabih, R. (2001) Fast Approximate Energy Minimization via Graph Cuts. *IEEE Trans. Pattern Anal. Mach. Intell.* **23**(11) 1222-1239.

[4] Shi, J. and Malik, J. (2000) Normalized Cuts and Image Segmentation. *IEEE Trans. Pattern Anal. Mach. Intell.* **22**(8) 888-905.

[5] Mumford, D. (2002) Pattern Theory: the Mathematics of Perception. *Proceedings of the ICM*. **1** 401-422.

[6] Mumford, D. and Shah, J. (1989) Optimal Approximations by Piecewise Smooth Functions and Associated Variational-Problems. *Communications on Pure and Applied Mathematics*. **42**(5) 577-685.

[7] Fowlkes, C., Belongie, S., Chung, F. and Malik, J. (2004) Spectral Grouping Using the Nyström Method. *IEEE Trans. Pattern Anal. Mach. Intell.* **26**(2) 214-225.

[8] Cour, T., Benezit, F. and Shi, J. (2005) Spectral Segmentation with Multiscale Graph Decomposition. Proceedings of the 2005 IEEE Computer Society Conference on Computer Vision and Pattern Recognition (CVPR'05) - Volume 2 - Volume 02, 1069213 IEEE Computer Society pp. 1124-1131.

[9] Eriksson, A., Olsson, C. and Kahl, F. (2011) Normalized Cuts Revisited: A Reformulation for Segmentation with Linear Grouping Constraints. *J. Math. Imaging Vis.* **39**(1) 45-61.

[10] Linli, X. (2009) Fast normalized cut with linear constraints. In Wenye, L. and Schuurmans, D. (eds).

[11] Catté, F., Lions, P.-L., Morel, J.-M. and Coll, T. (1992) Image selective smoothing and edge detection by nonlinear diffusion. *SIAM J. Numer. Anal.* **29**(1) 182-193.

[12] Perona, P. and Malik, J. (1990) Scale-Space and Edge-Detection Using Anisotropic Diffusion. *Ieee Transactions on Pattern Analysis and Machine Intelligence*. **12**(7) 629-639.

[13] Buades, A., Coll, B. and Morel, J. M. (2005) A Review of Image Denoising Algorithms, with a New One. *Multiscale Modeling & Simulation*. **4**(2) 490-530.

[14] Tomasi, C. and Manduchi, R. (1998) Bilateral Filtering for Gray and Color Images. Proceedings of the Sixth International Conference on Computer Vision, 939190 IEEE Computer Society pp. 839.

[15] Elad, M. (2002) On the origin of the bilateral filter and ways to improve it.

[16] Durand, F. and Dorsey, J. (2002) Fast bilateral filtering for the display of high-dynamic-range images. *ACM Trans. Graph.* **21**(3) 257-266.

[17] Eisemann, E. and Durand, F. (2004) Flash photography enhancement via intrinsic relighting. *ACM Trans. Graph.* **23**(3) 673-678.

[18] Petschnigg, G., Szeliski, R., Agrawala, M., Cohen, M., Hoppe, H. and Toyama, K. (2004) Digital photography with flash and no-flash image pairs. *ACM Trans. Graph.* **23**(3) 664-672.

[19] Winnemöller, H., Olsen, S. C. and Gooch, B. (2006) Real-time video abstraction. ACM SIGGRAPH 2006 Papers, 1142018 ACM pp. 1221-1226.

[20] Paris, S. and Durand, F. (2009) A Fast Approximation of the Bilateral Filter Using a Signal Processing Approach. *Int. J. Comput. Vision*. **81**(1) 24-52.

[21] Paris, S., Kornprobst, P., Tumblin, J. and Durand, F. (2008) A gentle introduction to bilateral filtering and its applications. ACM SIGGRAPH 2008 classes, 1401134 ACM pp. 1-50.

[22] Weiss, B. (2006) Fast median and bilateral filtering. *ACM Trans. Graph.* **25**(3) 519-526.

[23] Fatih, P. (2008) Constant time O(1) bilateral filtering.

[24] Yang, Q., Tan, K. H. and Ahuja, N. (2009) Real-time O(1) bilateral filtering. CVPR. IEEE, Miami, FL.

[25] Gastal, E. S. L. and Oliveira, M. M. (2012) Adaptive manifolds

for real-time high-dimensional filtering. *ACM Trans. Graph.* **31**(4) 1-13.

[26] Gastal, E. S. L. and Oliveira, M. M. (2011) Domain transform for edge-aware image and video processing. *ACM Trans. Graph.* **30**(4) 1-12.

[27] Kass, M. and Solomon, J. (2010) Smoothed local histogram filters. *ACM Trans. Graph.* **29**(4) 1-10.

[28] Lafferty, J. D., McCallum, A. and Pereira, F. C. N. (2001) Conditional Random Fields: Probabilistic Models for Segmenting and Labeling Sequence Data. Proceedings of the Eighteenth International Conference on Machine Learning, 655813 Morgan Kaufmann Publishers Inc. pp. 282-289.

[29] Belkin, M. and Niyogi, P. (2003) Laplacian Eigenmaps for dimensionality reduction and data representation. *Neural Comput.* **15**(6) 1373-1396.

[30] Malik, J., Belongie, S., Leung, T. and Shi, J. B. (2001) Contour and texture analysis for image segmentation. *International Journal of Computer Vision*. **43**(1) 7-27.

[31] Golub, G. H. and Loan, C. F. V. (1996) Matrix Computations. *The Johns Hopkins University Press*.

[32] Weickert, J. (1998) Anisotropic Diffusion in Image Processing. *ECMI Series, Teubner-Verlag*